\documentclass[letterpaper, 10 pt, conference]{ieeeconf}  

\IEEEoverridecommandlockouts                              
\usepackage{cite}
\usepackage{amsmath,amssymb,amsfonts}
\usepackage{algorithmic}
\usepackage{graphicx}
\usepackage{textcomp}
\usepackage{xcolor}
\def\BibTeX{{\rm B\kern-.05em{\sc i\kern-.025em b}\kern-.08em
    T\kern-.1667em\lower.7ex\hbox{E}\kern-.125emX}}

\usepackage{hyperref}
\hypersetup{
  pdftitle={ACROSS: A Deformation-Based Cross-Modal Representation for Robotic Tactile Perception},
  pdfauthor={Wadhah Zai El Amri, Malte Kuhlmann, Nicolás Navarro-Guerrero},
  pdfsubject={Robotics (cs.RO); Artificial Intelligence (cs.AI),},%
  pdfkeywords={BioTac, DIGIT, Tactile sensors, Transfer learning},%
  colorlinks=true,
  linktoc=all,
  linkcolor=darkgray,
  citecolor=darkgray,
  urlcolor=black,
}
\usepackage[protrusion=true,expansion=true]{microtype}
\usepackage{textcomp}
\usepackage{multirow}
\usepackage{blindtext}
\usepackage{footmisc}
\usepackage{xcolor}

\makeatletter
\long\def\@makecaption#1#2{%
    \ifx\@captype\@IEEEtablestring%
    \begin{center}%
        \setlength{\baselineskip}{0.7\baselineskip} 
        {\footnotesize #1}\\{\footnotesize\scshape #2}%
    \end{center}%
    \@IEEEtablecaptionsepspace%
    \else
    \@IEEEfigurecaptionsepspace%
    \setbox\@tempboxa\hbox{\footnotesize #1.~~ #2}%
    \ifdim \wd\@tempboxa >\hsize%
    \setbox\@tempboxa\hbox{\footnotesize #1.~~ }%
    \parbox[t]{\hsize}{\footnotesize \noindent\unhbox\@tempboxa#2}%
    \else%
    \ifcenterfigcaptions \hbox to\hsize{\footnotesize\hfil\box\@tempboxa\hfil}%
    \else \hbox to\hsize{\footnotesize\box\@tempboxa\hfil}%
    \fi\fi\fi}
\makeatother

\begin{document}

\title{\LARGE \bf
ACROSS: A Deformation-Based Cross-Modal Representation for Robotic Tactile Perception}

\author{Wadhah Zai El Amri$^{1}$, Malte Kuhlmann$^{1}$ and Nicolás Navarro-Guerrero$^{1}$
\thanks{$^{1}$\href{https://www.l3s.de/}{L3S Research Center}, Leibniz Universität Hannover, Hanover, Germany
        {\tt\small \{\href{mailto:wadhah.zai@l3s.de}{wadhah.zai}, \href{mailto:malte.kuhlmann@l3s.de}{malte.kuhlmann}, \href{mailto:nicolas.navarro@l3s.de}{nicolas.navarro}\}@l3s.de }}
}

\maketitle

\begin{abstract}
Tactile perception is essential for human interaction with the environment and is becoming increasingly crucial in robotics. Tactile sensors like the BioTac mimic human fingertips and provide detailed interaction data. Despite its utility in applications like slip detection and object identification, this sensor is now deprecated, making many valuable datasets obsolete. However, recreating similar datasets with newer sensor technologies is both tedious and time-consuming. Therefore, adapting these existing datasets for use with new setups and modalities is crucial. In response, we introduce ACROSS, a novel framework for translating data between tactile sensors by exploiting sensor deformation information. We demonstrate the approach by translating BioTac signals into the DIGIT sensor. Our framework consists of first converting the input signals into 3D deformation meshes. We then transition from the 3D deformation mesh of one sensor to the mesh of another, and finally convert the generated 3D deformation mesh into the corresponding output space. We demonstrate our approach to the most challenging problem of going from a low-dimensional tactile representation to a high-dimensional one. In particular, we transfer the tactile signals of a BioTac sensor to DIGIT tactile images.  Our approach enables the continued use of valuable datasets and data exchange between groups with different setups.
\end{abstract}

\section{Introduction}
Tactile feedback is gaining significant attention in robotics \cite{Dahiya2010Tactile, Navarro-Guerrero2023VisuoHaptic}. 
Tactile sensors leverage various information modalities, come in diverse shapes and sizes, and are implemented in various technologies. This diversity makes the exchange of data and trained models challenging. Moreover, as sensor technology improves, datasets become obsolete. 
For instance, BioTac by SynTouch was a high-end tactile sensor designed like a human fingertip. It has an elastomer covering a rigid core filled with an incompressible conductive fluid. The sensor outputs voltage readings from 19 internal electrodes, capturing changes in the fluid.  These readings are processed as time-series signal data~\cite{Wettels2007Biomimetic,Wettels2014Multimodal,ZaiElAmri2024Optimizing}. The BioTac has been proven useful in various applications such as detecting object slips and the direction of slips~\cite{Garcia-Garcia2019TactileGCN, Zapata-Impata2019Learning} or identifying objects~\cite{Xu2013Tactile}. However, this sensor is now deprecated. Consequently, many influential datasets, such as the BioTac SP direction of slip dataset \cite{Zapata-Impata2019Learning}, the BioTac SP grasp stability dataset \cite{Garcia-Garcia2019TactileGCN} or the BioTac 2P grasp stability dataset \cite{Chebotar2016BiGS}, are now obsolete. These datasets capture sensor outputs, specifically BioTac signals, recorded while the sensors are mounted on robotic hands that grasp various objects under different conditions. The data is then used, for instance, to evaluate the stability of the grasps. Despite their obsolescence, such datasets remain important, as grasp stability and slip detection continue to be an active field of research~\cite{Gong2023TactileBased}.


Furthermore, designing and collecting similar datasets is a time-consuming and complex task. It requires careful consideration of various factors, such as the choice of sensors and their resolution, the data collection methods, and the labeling process, among other requirements.

Hence, there is a need to convert useful datasets into formats compatible with newer sensor modalities, even if they involve different robotic or sensor configurations. This data conversion would allow researchers to leverage intrinsic information still relevant to specific tasks, while also saving time and resources by avoiding the need to collect entirely new datasets.

To this end, we propose ACROSS, a versatile approach for transferring tactile data between sensors of varying resolutions, including low-to-high, high-to-high, and high-to-low resolution transfers~\cite{ZaiElAmri2024Transferring}. We demonstrate the effectiveness of ACROSS by converting low-resolution tactile time-series data from a BioTac sensor into a high-resolution vision-based DIGIT sensor~\cite{Lambeta2020DIGIT}. 
Our method enables the utilization of existing datasets gathered with outdated sensors, avoiding the tedious process of gathering data from scratch. Moreover, it facilitates a transition between two intrinsically distinct tactile sensor modalities, e.g., signal data, to visual representations. 

While highly accurate simulations could generate realistic sensor data, they are insufficient for direct cross-sensor transfer, as they do not intrinsically provide a way to relate the outputs of different sensors or bridge the gap between their modalities. Current physics simulation environments, such as Isaac Gym~\cite{Isaac}, primarily model sensors' deformation during object interaction but do not provide the actual sensor-specific outputs, such as BioTac sensor signals or DIGIT sensor images. Moreover, these environments do not provide a way to map deformations between different sensors, even when both are simulated within the same environment. ACROSS addresses this gap by enabling the transfer of sensor outputs, even across different simulation environments, making it a crucial framework for data transfer beyond what direct simulation alone can achieve.

Additionally, we provide an openly available dataset comprising over 155K unique 3D mesh deformation pairs from interactions involving BioTac and DIGIT sensors. This dataset includes various types of indenters and the force exerted on each sensor.
The source code, dataset, and neural network checkpoints can be found on our website: \href{https://wzaielamri.github.io/publication/across}{\textcolor{gray}{https://wzaielamri.github.io/publication/across}}.

\section{Related Work}
Tactile sensors can capture the same deformation of an object, but they may represent this deformation differently depending on the type of sensor used~\cite{Dahiya2013Robotic}. For instance, the elastomer's deformations can be represented as either time-series signals or images, depending on the modality of the sensor. Therefore, our proposed approach focuses on transferring the encoded information and knowledge at the deformation level rather than at the output level, which varies between sensors. Although some research attempted to transfer between modalities, for instance, Lee et al.~\cite{Lee2019Touching} developed a framework to generate tactile images from the GelSight sensor using digital camera images of various cloth materials and vice versa, or the ViTac dataset~\cite{Luo2018ViTac}, used to train the networks, includes labeled images from the GelSight sensor and a digital camera of 100 fabric pieces. Their framework utilizes two separate cycleGANs for the bidirectional transfer. Unfortunately, these approaches require labeled data from both sensor types. Moreover, the modality of the source and target sensors is the same, i.e., vision.

Tatiya et al.~\cite{Tatiya2020Framework} introduced a framework for transferring knowledge across sensor modalities, allowing robots to handle sensor failures or operate with different sensor configurations. Their method leverages a variational encoder-decoder network (VED) to map sensory observations from one modality, such as vibration, to another, such as haptic, by learning a shared feature space between them. However, the approach similarly requires end-to-end labeled data, which can be resource-intensive. Additionally, the framework assumes the same set of objects is used across experiments, potentially limiting its generalization to novel objects.

In contrast, our approach does not require end-to-end labeled sensor outputs and can generalize to any contact form, force, or orientation. Furthermore, we facilitate transfers between distinct modalities and sensors with different morphologies and sizes.
Our framework starts by converting the input signals into 3D deformation meshes. Next, we transition from the 3D deformation mesh of one sensor to that of another, and finally, we translate the generated 3D deformation mesh into the corresponding output space. Unlike prior research, we address the challenging task of converting a low-dimensional tactile representation into a high-dimensional one. Specifically, we transfer tactile signals from a BioTac sensor to DIGIT tactile images.

Our approach is inspired by Narang et al.~\cite{Narang2021Interpreting,Narang2021SimtoReal}, who introduced a framework using a finite element method (FEM) model to simulate the deformation of the BioTac sensor interacting with different indenters. Two variational autoencoders (VAE) were used. The first, a vanilla VAE with linear layers, reconstructs the BioTac signals, while the second, a convolutional mesh autoencoder (CoMA), reconstructs the mesh deformation. CoMA network employs fast localized spectral filtering, i.e., Chebyshev filters alongside hierarchical pooling operations. These operations are adapted for 3D meshes by computing the spectral information of the mesh graph using Fourier transformation and then applying Chebyshev filters for localized convolutional operations~\cite{Defferrard2016Convolutional,Ranjan2018Generating}. Both VAEs are subsequently `frozen', and an MLP is trained to project the latent vector from one modality to another using labeled data pairs. 
In our work, we adopt Narang et al.'s~\cite{Narang2021Interpreting} approach, due to its performance, to generate BioTac deformation meshes from the sensor input signals. 

Zhu et al.~\cite{Zhu2022Learning} applied a similar idea to synthesize the volumetric mesh of a vision-based tactile sensor, GelSlim~\cite{Taylor2022GelSlim}. Two separate VAEs were employed. The first VAE was trained to reconstruct the tactile images captured by GelSlim, ensuring that the network could accurately capture the visual features of the deformed elastomer. The second VAE was dedicated to reconstructing the volumetric mesh from the tactile imprints, capturing the 3D structure of the deformation. Zhu et al. introduced a self-supervised adaptation method that leverages a differentiable renderer to generate synthetic meshes to refine their approach further. This technique improved the  encoder's performance, which embeds the image into the latent space, and the projection Multilayer Perceptron (MLP), which maps between the two different latent representation spaces of the images and the volumetric mesh. By rendering the generated mesh using the differentiable renderer and comparing it to the observed tactile imprint, the system could compute gradients and backpropagate errors, thereby refining the model through further training. This combination of self-supervision and differentiable rendering allowed the system to bridge the gap between simulation and real-world tactile data and generate accurate mesh deformations from tactile images.

In contrast, other researchers have focused on creating images derived from the physical interactions of vision-based sensors. For instance, Wang et al.~\cite{Wang2022TACTO} provided TACTO, a simulation of vision-based sensors, which uses Pyrender~\cite{mmatl} and normal forces to derive gel pad deformations and generate then the corresponding sensor images. This simulation is primarily demonstrated for the  DIGIT sensor~\cite{Lambeta2020DIGIT}. A vision-based sensor that consists of an elastomeric gel pad that reflects light emitted by a series of internal LEDs, enabling an internal camera to record the deformation of the gel membrane. Another solution, Taxim~\cite{Si2022Taxim}, predicts the output of image-based tactile sensors using example-based photometric stereo methods. It utilizes optical reflection functions to interpret the gel pad's illumination while interacting with objects. The core concept involves simulating the sensor's optical output using a polynomial table based on a second-order polynomial function. This function approximates the non-linearity of light in vision-based tactile sensors. Additionally, this approach enables the calibration and adjustment of these polynomial coefficients with real sensor data and allows for the simulation of marker motion fields on the gel surface. Given Taxim's superior performance over other state-of-the-art methods, we incorporate it into our framework to generate DIGIT images from 3D deformation meshes.

\section{Implementation}\label{sec:implementation}
\begin{figure}[tbp]
  \centering
  \includegraphics[width=1.0\columnwidth]{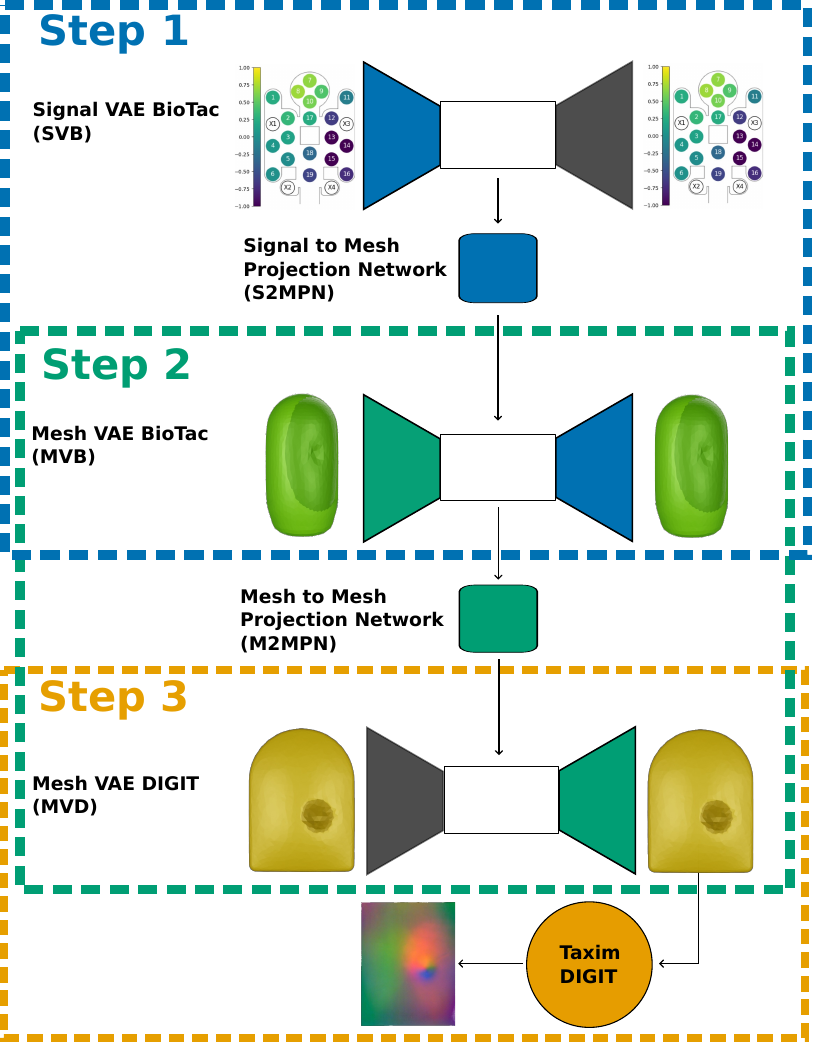}
  \caption{An example of the ACROSS framework applied to translate BioTac signals into DIGIT images. Step 1: Convert BioTac input signals to BioTac surface deformation. Step 2: Convert BioTac surface mesh deformation to DIGIT surface mesh deformation. Step 3: Generate DIGIT's output from the surface mesh deformation.}
  \label{fig:full_pipeline_networks}
\end{figure}

Transferring between different modalities poses challenges due to data representations and encoding variations. We propose a three-step solution to address these issues, depicted in Figure~\ref{fig:full_pipeline_networks}. Step I: We initially predict the BioTac surface deformation from the BioTac input signals. Step II: We convert the BioTac surface mesh deformation to DIGIT surface mesh deformation since the physical interaction of both sensors can be modeled by a mesh deformation independently of the sensor output modality. Step III: We generate the DIGIT sensor image from the converted deformation. 

\subsection*{Step I: Predicting BioTac Mesh Deformation}
We adopt a similar methodology to that proposed by Narang et al.~\cite{Narang2021SimtoReal}. We train a disentangled variational autoencoder ($\beta$-VAE)~\cite{Higgins2017betaVAE} to reconstruct the BioTac sensor outputs. This network is denoted as Signal VAE BioTac ({\textit{SVB}}).
To train the network, we use a curated dataset that combines two publicly available BioTac signal datasets, i.e., Narang et al.~\cite{Narang2021SimtoReal} and Ruppel et al.~\cite{Ruppel2019Simulation}. This combination allows us to effectively augment our data, increasing the diversity and amount of data available for training. By augmenting our dataset, we empirically reduced the overall loss and helped the network learn more robust features by exposing it to a broader range of examples.

We also train another $\beta$-VAE to reconstruct the 3D mesh deformation of the BioTac sensor, denoted as Mesh VAE BioTac ({\textit{MVB}}). To model these deformations and collect the dataset used for this task, we employ the validated Isaac Gym BioTac FEM simulation~\cite{Narang2021SimtoReal}. 

Next, we train an MLP network to map between the latent vectors of the \textit{SVB} and the \textit{MVB} network, referred to as Signal to Mesh Projection Network ({\textit{S2MPN}}). For this latent space mapping, we use the publicly available dataset collected by Narang et al.~\cite{Narang2021SimtoReal}. This dataset comprises data pairs: BioTac electrodes outputs and mesh deformations, resulting from interactions with nine different indenters.

\subsection*{Step II: Modeling of Mesh Deformation}
In this step, we train a third $\beta$-VAE~\cite{Higgins2017betaVAE}, with the same architecture used for the \textit{MVB} network, this time to reconstruct the DIGIT 3D deformations. This network is denoted as Mesh VAE DIGIT ({\textit{MVD}). The data used to train this network are also collected with Isaac Gym through simulating physical interactions with the DIGIT sensor and recording the corresponding 3D deformations. Given that the sensor's elastomer is primarily composed of Smooth-On Solaris silicone~\cite{Lambeta2020DIGIT}, we configure the FEM soft-body hyperparameters accordingly: an elasticity modulus of 539 kPa and a Poisson's ratio of 0.499~\cite{Schoenborn2023FluidStructure}. Furthermore, we set the dynamic friction coefficient between the DIGIT sensor elastomer and the indenters to 0.78~\cite{Narang2021SimtoReal}. 

Afterward, we train an MLP network to map the latent space of the already trained \textit{MVB} encoder network in Step I to the latent space of the trained \textit{MVD} encoder network, and we denote this network as Mesh to Mesh Projection Network (\textit{M2MPN}). To train the \textit{M2MPN}, we collected unique paired mesh deformations for both BioTac and DIGIT using the Isaac Gym simulator. Details about the dataset collection procedures follow in Section~\ref{sec:dataset}.

\subsection*{Step III: From Surface Deformation to DIGIT's Output}
We adapt the simulation model \textit{Taxim}~\cite{Si2022Taxim} to simulate DIGIT images. Originally, Taxim calculated a height map of the gel pad using object point clouds to estimate the corresponding DIGIT image. We adjust this approach using the deformation mesh instead of the point cloud. Using Pyrender~\cite{mmatl}, we estimate the height map that the sensor's internal camera would capture for each deformation mesh. Subsequently, we generate the corresponding DIGIT image by applying this height map and Taxim's polynomial coefficients. Later, we improve the synthetic image by applying a pyramid Gaussian blur to remove its artifacts and make it more realistic. We describe this in detail in Section~\ref{sec:NetDescrip}.


\section{Experiments and Results}\label{sec:experiments_results}
This section introduces the datasets used in our framework, followed by a detailed description of our network architecture. Finally, it presents the results of our approach.

\subsection{Datasets Description}\label{sec:dataset}
To train the \textit{SVB} network, we curated an unlabeled real BioTac signal dataset by normalizing each input channel separately and merging two existing datasets~\cite{Narang2021SimtoReal,Ruppel2019Simulation} to augment our data and improve the network’s generalization and performance on unseen data. Each data vector comprises 19 electrode values, normalized and adjusted to fall within the range of [-1, 1]. The dataset provided by Ruppel et al.~\cite{Ruppel2019Simulation} contains an error that increased throughout data gathering. This error is attributed to a rise in temperature during the data collection process, which affects the properties of the fluid and causes an output drift~\cite{ZaiElAmri2024Optimizing}. To address this, we fitted a linear function for each electrode by utilizing gradient descent to minimize the difference between each non-contact timestep and the linear function. We then used the linear function to shift the values towards the default values of the electrodes. 
Default values represent the sensor readings at rest, i.e., no touches are recorded. 
Furthermore, most non-contact data was excluded to ensure a balanced dataset. The combined dataset yielded almost 250K unique BioTac signal data points.



We then collect a BioTac deformation dataset consisting of approximately 860 unique indentations trajectories. Each trajectory includes 20 different 3D mesh deformations, with depths ranging from $0.1\:mm$ to $2\:mm$ in $0.1\:mm$ increments. We set the maximum indentation to $2\:mm$, a value chosen to align with later simulations involving the DIGIT sensor. Given that the DIGIT has thinner sides, $2\:mm$ is a suitable maximum indentation for these simulations. Furthermore, each trajectory is replicated using the nine indenter types shown in Figure~\ref{fig:indenters_images} to have a wide variety of touches. This results in an overall dataset of approximately 155K unique 3D deformation meshes. We use the collected dataset to train the \textit{MVB} network.

\begin{figure}[tbp]
  \centering
  \includegraphics[width=0.7\columnwidth]{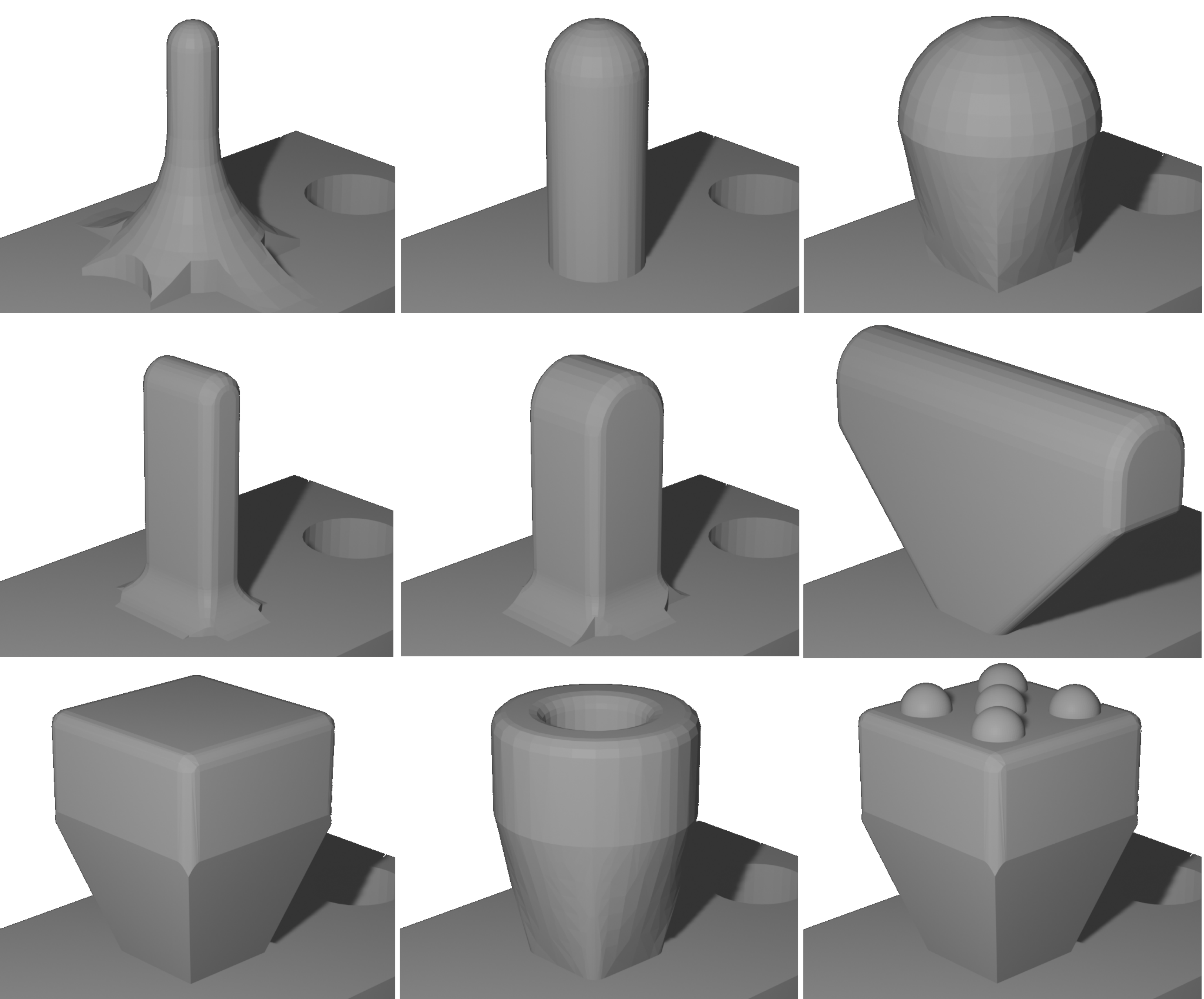}
  \caption{The nine indenters used to collect the BioTac-DIGIT deformation dataset.}
  \label{fig:indenters_images}
\end{figure}

Since the \textit{S2MPN} requires labeled data to train, we re-simulate the BioTac mesh deformations corresponding to the real signals in Narang et al.'s~\cite{Narang2021SimtoReal} dataset, following their description. The dataset was then filtered to ensure consistency with our collected BioTac deformation dataset. Only data points within a maximum depth of $2\;mm$ were selected, resulting in a dataset of almost 9K unique data points with the corresponding labeled BioTac signals. The data was divided into distinct trajectories to ensure an unbiased split. We then randomly selected $15\%$ of those trajectories as our test set. The remaining data points were randomly split into training and validation sets, as shown in Table~\ref{tab:datasets_splits}.

In Step II, we collected a DIGIT mesh deformation dataset comprising approximately 155K unique 3D DIGIT mesh deformations, following the same trajectories as the BioTac mesh dataset.
To ensure alignment between the two mesh datasets, we maintained consistent force, angle, and position parameters for each interaction pair since labeled and matching BioTac DIGIT meshes are required for training the \textit{M2MPN}. In order to address the difficulty in representing side touches arising from the differing shapes of the DIGIT and BioTac sensors, we rotate and translate the BioTac sensor on its axis within its horizontal plane, as depicted in Figure~\ref{fig:biotac_unfolding_digit}. This transformation mimics the unfolding of the BioTac elastomer to align with and cover the flat surface of the DIGIT sensor. This solution ensures that forces and deformations resulting from side touches correspond with the other sensor.

\begin{figure}[tbp]
  \centering
  \includegraphics[width=0.75\columnwidth]{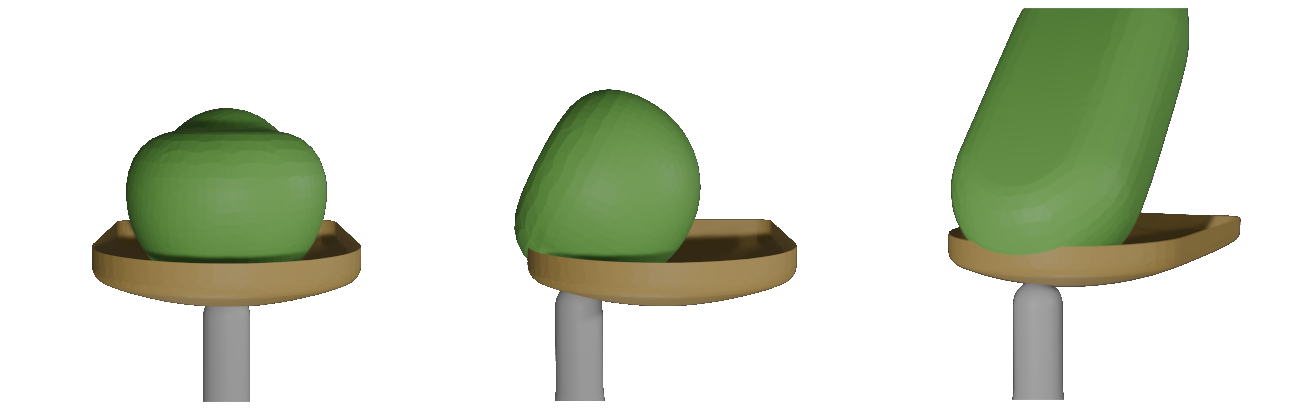}
  \caption{Transferred BioTac sensor (green) to align it with the DIGIT sensor surface (gold), in order to collect paired 3D deformation meshes.}
  \label{fig:biotac_unfolding_digit}
\end{figure}

\begin{table}[tbp]
\centering
\caption{Size of the datasets used in our framework. The number between parenthesis presents the size of the input/ output of the data. The same datasets are used for $^1$ and $^2$, and the splits are identical.}\label{tab:datasets_splits}
\resizebox{\columnwidth}{!}{
\setlength\tabcolsep{4pt} 
{\renewcommand{\arraystretch}{1.5}
\begin{tabular}{@{}l|l|c|c|c@{}}
\hline
\textbf{Networks}       & \textbf{Datasets}                                                                         & \textbf{Train} & \textbf{Validation} & \textbf{Test} \\ \hline\hline
\textbf{SVB}     & BioTac Signals (19)                                                                       & \hfill 196397         & \hfill 27218               & \hfill 24551          \\ \hline
\textbf{S2MPN} & \begin{tabular}[c]{@{}l@{}}BioTac Signals (19)\\BioTac Meshes (4246x3)\end{tabular}   & \hfill 5841           & \hfill 1409               & \hfill 1417          \\ \hline
\textbf{MVB}       & BioTac Meshes (4246x3)$^1$                                                                    & \hfill 122924         & \hfill 15560               & \hfill 17121         \\ \hline
\textbf{MVD}        & DIGIT Meshes (6103x3)$^2$                                                                     & \hfill 122924         & \hfill 15560               & \hfill 17121         \\ \hline
\textbf{M2MPN}   & \begin{tabular}[c]{@{}l@{}}BioTac Meshes (4246x3)$^1$\\DIGIT Meshes (6103x3)$^2$ \end{tabular} & \hfill 122924         & \hfill 15560               & \hfill 17121         \\ \hline
\end{tabular}%
}
}
\end{table}
\renewcommand{\arraystretch}{\originalarraystretch}

\subsection{Network Descriptions}~\label{sec:NetDescrip}
The encoder of our \textit{SVB} network comprises five layers. The first three are linear layers with sizes $[256, 128, 64]$ and two parallel linear layers which predict $\mu$ and $log(\sigma^2)$, each with a size of $8$. The decoder is composed of four linear layers with sizes $[64, 128, 256, 19]$. Each layer uses a ReLU activation function. The network is trained to minimize the following function:
\begin{equation}\label{eq:loss_signal_vae}
    \ell_S=\operatorname{MSE}(S-\hat{S})+\beta_S \operatorname{KL}\left(f\left(z_S \mid S\right) \| \mathcal{N}(0,1)\right),
\end{equation}
where $f$ is the encoder of the SVB network, $\beta_S$ is the weight of the KL divergence loss, and is equal to $0.005$. $z_S$ is the sampled latent vector given the input $S$. The mean-squared error (MSE) is calculated between the normalized predicted and ground-truth signal. The network uses the Adam optimizer with a learning rate equal to $0.0001$. 
The hyperparameters for all our proposed network architectures are empirically determined.

The \textit{S2MPN} has four linear layers with sizes $[512, 128, 256, 256]$. Each linear layer is followed by an ELU activation function and a dropout layer with dropout rates equal to $[0.4, 0.3, 0.2, 0.5]$. It is optimized using Adam with a learning rate of $0.0005$ and using the following MSE loss function:
\begin{equation} \label{eq:loss_signal_to_mesh_projection}
    \ell_{SMP}=\operatorname{MSE}(z_{MB}-\hat{z}_{MB}),
\end{equation}
where $z_{MB}$ represents the predicted BioTac latent space and $\hat{z}_{MB}$ is the target BioTac latent space.

Both our 3D mesh reconstruction VAEs, i.e., \textit{MVB} and \textit{MVD}, are built upon graph convolutional mesh autoencoders (CoMA). 
Both networks have identical architectures. The encoder is composed of four graph convolution layers with sizes $[16, 16, 16, 32]$ and a kernel size of 6, each followed by a downsampling layer with a factor of 2. Next, a linear layer with a size of 512 is applied, followed by two parallel linear layers with sizes $[128, 128]$ to represent $\mu$ and $log(\sigma^2)$, used for the latent space of the VAE. 
The networks are optimized using Adam, with an initial learning rate of $0.001$ and a learning rate decay after each epoch equal to $0.99$. The following cost function is minimized:
\begin{equation}\label{eq:loss_mesh_vae}
    \ell_M=\operatorname{MSE}(M-\hat{M})+\beta_M \operatorname{KL}\left(g\left(z_M \mid M\right) \| \mathcal{N}(0,1)\right),
\end{equation} 
where $g$ is the encoder of the corresponding mesh VAE, $\beta_M$ is the weight of the KL divergence loss, and is equal to $0.005$ for both BioTac and DIGIT mesh VAEs. $z_M$ is the sampled latent vector given the input $M$. The mean-squared error (MSE) is calculated between the input and predicted normalized mesh, averaged over the entire batch and all 3D vertices. All networks are trained for a maximum of 300 epochs, with early stopping to prevent overfitting.

Finally, our \textit{M2MPN} consists of four linear layers with sizes $[512, 1024, 1024, 256]$. Each linear layer is followed by a dropout layer with $[0.2, 0.4, 0.0, 0.0]$ dropout rates, and an ELU activation function. The learning rate is set to 0.001. Early stopping is also employed, and the minimized loss function is defined as follows:
\begin{equation} \label{eq:loss_mesh_projection}
    \ell_{MMP}=\operatorname{MSE}(z_{MD}-\hat{z}_{MD}),
\end{equation}
where $z_{MD}$ represents the predicted DIGIT latent space vector and $\hat{z}_{MD}$ is the target latent space vector.

In the final step of this pipeline, we use Pyrender~\cite{mmatl} to obtain the height map of each DIGIT mesh deformation. The Taxim images generated from these height maps contain some artifacts due to the resolution of the gel pad mesh. To address this issue, we apply an additional pyramid Gaussian blur to the entire generated image, including the contact region, using kernel sizes of [51, 21, 11, 5]. This differs from the original Taxim, which does not apply Gaussian blur to the contact region. Figure~\ref{fig:taxim_gaussianBlur} illustrates the artifacts and the improvements achieved after applying the additional pyramid Gaussian blur.

\begin{figure}[tbp]
  \centering
  \includegraphics[width=0.75\columnwidth]{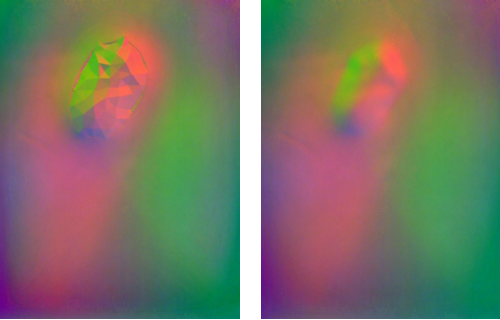}
  \caption{Comparison of artifacts in the generated image before (left) and after (right) applying the additional pyramid Gaussian blur.}
  \label{fig:taxim_gaussianBlur}
\end{figure}

\subsection{Converting Real Data}
In this subsection, we assess the performance of our trained networks in converting real-world data. We calculate the root-mean-square error (RMSE) averaged on all unseen test data for all our trained networks to quantify this. The RMSE for the \textit{SVB} is measured between the ground truth normalized electrode values and the network prediction, and is equal to $0.060$ with a standard deviation of $0.034$. 
When testing both projector networks, i.e., \textit{S2MPN} and \textit{M2MPN}, we generate the mesh from the predicted latent vector using the frozen trained decoder of the corresponding VAE network, and we measure the average RMSE in micrometers ($\mu m$) for all vertices between the ground-truth mesh and predicted mesh.
Additionally, we calculate the RMSE averaged only for the vertices within the deformation region. The deformation region is defined as comprising all vertices that deviate by $10\:\mu m$  or more from the original mesh without indentations. 
Furthermore, we measure the Euclidean distance between the predicted and target meshes in micrometers ($\mu m$) for all vertices and those specifically within the deformation region. The results of S2MPN, MVB, MVD and M2MPN are reported in Table~\ref{tab:results}.

\begin{table}[tbp]

\caption{Root-mean-square error (RMSE) and Euclidean distance (Euc. Dist.) measured between the prediction and the ground-truth outputs for all samples in the test set. For all networks, the RMSE and Euc. Dist. are reported in $\mu m$. Values are averaged over all the mesh vertices and on only vertices in the deformation region.}
\label{tab:results}
\resizebox{\columnwidth}{!}{%
\setlength\tabcolsep{3pt} 
\centering
\renewcommand{\arraystretch}{1.5}%
{
\begin{tabular}{@{}l|cc|cc@{}}
\hline
\multirow{2}{*}{\textbf{Networks}} & \multicolumn{1}{c|}{\textbf{RMSE}} & \textbf{Euc. Dist.} & \multicolumn{1}{c|}{\textbf{RMSE}} & \textbf{Euc. Dist.} \\ \cline{2-5} 
 & \multicolumn{2}{c|}{\textbf{All Vertices}} & \multicolumn{2}{c@{}}{\textbf{Deformation Region}} \\ \hline  \hline
\textbf{S2MPN} & \multicolumn{1}{c|}{\hspace{1.35mm}78.21 (41.88)} & \hspace{1.35mm}85.00 (49.80) & \multicolumn{1}{c|}{\hspace{1.35mm}94.07 (48.32)} & 121.02 (62.31) \\ \hline
\textbf{MVB} & \multicolumn{1}{c|}{\hspace{1.35mm}12.28 \hspace{1.35mm}(4.75)} & \hspace{1.35mm}13.90 \hspace{1.35mm}(5.15) & \multicolumn{1}{c|}{\hspace{1.35mm}16.03 \hspace{1.35mm}(4.29)} & \hspace{1.35mm}21.26 \hspace{1.35mm}(4.92) \\ \hline
\textbf{MVD} & \multicolumn{1}{c|}{\hspace{1.35mm}\hspace{1.35mm}9.28 \hspace{1.35mm}(4.20)} & \hspace{1.35mm}10.13 \hspace{1.35mm}(3.98) & \multicolumn{1}{c|}{\hspace{1.35mm}12.43 \hspace{1.35mm}(3.93)} & \hspace{1.35mm}14.60 \hspace{1.35mm}(4.15) \\ \hline
\textbf{M2MPN} & \multicolumn{1}{c|}{\hspace{1.35mm}21.57 (19.08)} & \hspace{1.35mm}18.68 (19.38) & \multicolumn{1}{c|}{\hspace{1.35mm}27.72 (20.07)} & \hspace{1.35mm}28.54 (21.90) \\ \hline
\end{tabular}
}
}
\end{table}
\renewcommand{\arraystretch}{\originalarraystretch}


According to Table~\ref{tab:results}, \textit{S2MPN} exhibits higher RMSE and Euclidean distance error than the other networks. This performance discrepancy is primarily attributed to the limited training set of paired examples. Additionally, the used dataset from Narang et al.~\cite{Narang2021Interpreting} includes misaligned signals and indenters positions that could not be corrected.
%


Further, we tested our entire framework on unseen real BioTac signal recordings from Narang et al.~\cite{Narang2021SimtoReal} and converted them to DIGIT output images. Figure~\ref{fig:results_samples} shows the qualitative results of five selected BioTac signals that we converted to DIGIT images using our framework.

\begin{figure}[tbp]
  \centering
  \includegraphics[width=0.95\columnwidth]{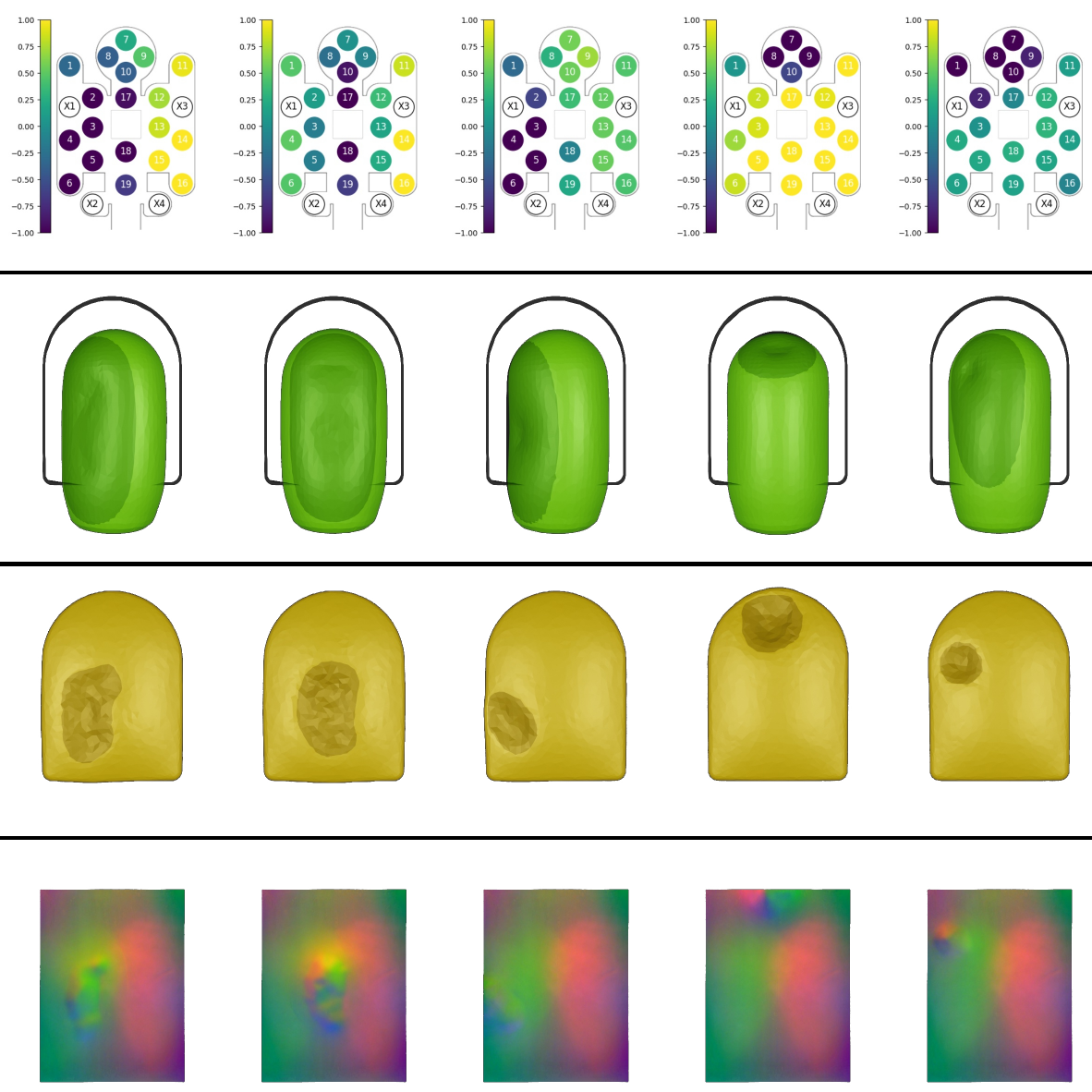}
  \caption{Converted samples. First row: Real electrode values. Second row: Ground-truth BioTac mesh deformations. The outer frame represents the ``unfolded'' BioTac surface. Third row: Converted DIGIT mesh deformations. Fourth row: DIGIT output images. The third and fourth rows were generated using the first row as input.}
  \label{fig:results_samples}
\end{figure}

The spatial positions and depth of the indentations are accurately preserved between the BioTac input signal and the DIGIT output image across all timestamps in the test set, and not only in those examples shown in Figure~\ref{fig:results_samples}. These accurate results can be verified in the video available on our website. However, the shapes of the indentations are partially reconstructed. 
This limitation arises due to the lower resolution of the BioTac sensor, which has only 19 electrodes. Compared to the higher-resolution DIGIT sensor output, this sensor cannot capture the detailed contours of indentations.

\section{Discussion and Conclusion}\label{sec:discussion_conculsion}
Our novel framework, ACROSS, represents a promising approach for re-utilizing datasets from deprecated sensors, and it enables the exchange of data across different setups. Despite differences in sensor modalities, we successfully demonstrated the framework's capability to accurately convert previously unseen BioTac sensor data into DIGIT output images, as evidenced by the qualitative examples provided. 

ACROSS consists of three steps: The first step involves converting the source sensor inputs into their corresponding 3D deformations. The second step involves mapping these source sensor deformation meshes to the target sensor deformation meshes. Finally, in the third step, we generate the output values based on the resulting meshes.

The core innovation of our framework lies in the 3D mesh deformation conversion between tactile sensors, which share an inherent similarity. We demonstrate this approach by transferring low-resolution inputs, i.e., BioTac signals, into high-resolution outputs, i.e., DIGIT images. Nevertheless, the lower resolution and differing format of the BioTac sensor compared to the DIGIT sensor may result in the loss of detail that a real DIGIT sensor would capture, such as the precise shape of indentations.

To this end, we offer a dataset featuring paired 3D mesh deformations from BioTac and DIGIT sensors, as well as a DIGIT FEM model for simulating the mesh deformations. However, the current framework has limitations. For instance, given the curvature of the sensor surfaces and shape mismatches, precise alignment of the sensors was essential during data collection to accurately capture side touches on both surfaces. In the current paper, we addressed this by calculating a transformation matrix to align the BioTac surface with the DIGIT surface, involving rotation and translation of the BioTac sensor within its horizontal plane. Hence, mapping physical deformations between sensors with different morphologies and features is inherently challenging. We aim to develop an alternative method for representing the data that does not require calculating transformation matrices for alignment.

The proposed approach relies on modeling the deformation of the sensor, and its effectiveness is closely tied to the accuracy of simulations. Thus, making a validated simulation of this deformation a necessary prerequisite. While this allows the framework to generalize to various tactile sensors, it fundamentally limits its applicability to sensors for which such simulations exist. For example, vibration-based sensors~\cite{JuinaQuilachamin2023Fingerprint}, which are considered for tactile sensing, do not undergo deformation in a way that can be meaningfully simulated within this framework. Similarly, other non-deformable sensing technologies, such as capacitive sensors~\cite{chiRecentProgressTechnologies2018}, may not fit within this approach.
 
Future work will focus on improving the signal-to-mesh model by expanding the training dataset and gathering more real-world data. We also plan to quantitatively assess the conversion quality by applying the framework to various tasks, i.e., classification tasks, and explore its adaptability by incorporating additional sensor modalities. Furthermore, by exploring alternative methods, we intend to refine the syntactic data generated by the Taxim algorithm, which currently lacks shadow information for the DIGIT sensor. We aim to improve the framework's robustness and ability to generalize effectively to other sensors.

\addtolength{\textheight}{-65mm} 

\bibliographystyle{IEEEtran}
\bibliography{references.bib}

\end{document}